# Path Planning of Cleaning Robot with Reinforcement Learning

Woohyeon Moon*
*Cho Chun Shik Graduate School of Mobility*
*Korea Advanced Institute of Science and Technology (KAIST)*
Daejeon, South Korea
moonstar@kaist.ac.kr

Bumgeun Park*
*Cho Chun Shik Graduate School of Mobility*
*Korea Advanced Institute of Science and Technology (KAIST)*
Daejeon, South Korea
j4t123@kaist.ac.kr

Sarvar Hussain Nengroo
*Cho Chun Shik Graduate School of Mobility*
*Korea Advanced Institute of Science and Technology (KAIST)*
Daejeon, South Korea
sarvar@kaist.ac.kr

Taeyoung Kim
*Cho Chun Shik Graduate School of Mobility*
*Korea Advanced Institute of Science and Technology (KAIST)*
Daejeon, South Korea
ngng9957@kaist.ac.kr

Dongsoo Har
*Cho Chun Shik Graduate School of Mobility*
*Korea Advanced Institute of Science and Technology (KAIST)*
Daejeon, South Korea
dshar@kaist.ac.kr

***Abstract*— Recently, as the demand for cleaning robots has steadily increased, therefore household electricity consumption is also increasing. To solve this electricity consumption issue, the problem of efficient path planning for cleaning robot has become important and many studies have been conducted. However, most of them are about moving along a simple path segment, not about the whole path to clean all places. As the emerging deep learning technique, reinforcement learning (RL) has been adopted for cleaning robot. However, the models for RL operate only in a specific cleaning environment, not the various cleaning environment. The problem is that the models have to retrain whenever the cleaning environment changes. To solve this problem, the proximal policy optimization (PPO) algorithm is combined with an efficient path planning that operates in various cleaning environments, using transfer learning (TL), detection nearest cleaned tile, reward shaping, and making elite set methods. The proposed method is validated with an ablation study and comparison with conventional methods such as random and zigzag. The experimental results demonstrate that the proposed method achieves improved training performance and increased convergence speed over the original PPO. And it also demonstrates that this proposed method is better performance than conventional methods (random, zigzag).**

## I. INTRODUCTION

Currently, robots are being used in many fields such as factories, restaurants, and hotels, thanks to the recent development of robotic technology. These robots have been used not only in the commercial environment but also in our living environment. At Consumer Electronics Show 2019, Samsung Electronics came into the spotlight by introducing care robots and home robots. Numerous companies around the world have introduced many home robots and among them cleaning robots are outstanding. The technical advancements of cleaning robot and increased usage of the cleaning robot drive the market for cleaning robot.

Lately, the cleaning robot is equipped with a system that uses artificial intelligence (AI) to address the challenge of cleaning a large area while taking into account the variables like the number of turns and the length of the trajectory. The AI has been instrumental in the usage of numerous robot applications, including mobile cleaning, helping elderly persons, driving underwater, operating aerial vehicles, and farming [1, 2]. In case of cleaning, the cleaning robot thoroughly cleans every accessible space across the entire room, considering the relevant map. Different algorithms and techniques have been employed to fulfill this task in dynamic situations [1, 2].

Adoption of the robot to replace human employee becomes important especially when risks exist at work [3]. Recently, robots have become common household items due to the rising human needs. According to semiconductor and electronics industry research, the market for cleaning robots is anticipated to reach USD 4.34 billion in 2023, with a compound annual growth rate of 16.21% between 2018 and 2023. Most of the studies on cleaning robot since 2002 have concentrated on lowering development costs and escalating work efficiency. To achieve this goal, numerous enhancements have been developed by using multiple sensors and well-designed path planning algorithms. For instance, a self-contained cleaning robot can be used to extend the lifespan and efficiency of the solar panel [4]. Another example could be finding the best path through a cluttered environment by improved particle swarm optimization and gravitational search algorithm [5].

In spite of the recent evolution of robotic technology, there are lingering questions about the efficiency of cleaning robots. One of the challenges in creating a high-quality cleaning robot is efficient path planning all around the working environment. Most cleaning robots are moving straightforwardly, and when obstacles such as a wall, chair, and table are located in front of the cleaning robot, they just change direction and go straight until the next obstacle appears. This method not only has the problem of unnecessary power consumption but also intercepts the path of moving people. To solve this problem, a few studies using RL algorithms have been conducted [6]. However, most of the models for RL are trained in a specific environment. In this case, the trained model does not work in a different environment, e.g., a different room. Therefore, learning is difficult even if the structure of the house is already known.

*Woohyeon Moon and Bumgeun Park contributed equally to this work.

To solve the problem of inefficient cleaning path [7-9], we proposed a model by using the PPO algorithm and combining techniques of transfer learning (TL), detection of nearest uncleaned tile (DNUT), reward shaping (RS), and making elite set (ES) with the PPO algorithm to improve the performance.

The main contributions of this work are summarized as follows.

1) The PPO algorithm, an on-policy algorithm among RL algorithms fit to optimization of the path of cleaning robots is combined with other supplementary techniques to provide improved performance of the cleaning robot.

2) Supplementary techniques of TL, DNUT, RS, and making ES are suggested to be combined with the PPO algorithm to improve the performance of learning.

3) Comparative performance analysis between the existing and the proposed method obtained from the PPO algorithm combined with the supplementary techniques is presented for a typical home environment.

## II. Background

In this section, the concepts and the mathematical models of RL, Value-based RL, Policy-based RL, Deep Q Network, PPO, and TL are presented.

### A. Reinforcement Learning

The basic concept of RL came from behavioral psychology [10] Activist psychologist Skinner created a 'Reinforcement Theory' by confirming that animals learn the relationship between behavior and its consequences. An agent such as an animal learns the correlation between behavior and the resulting good reward, which has not been learned before. Computers that learn by themselves through RL are often called 'agents'. These agents learn by receiving rewards while taking action in a certain environment. Recently, surprising research results related to this theory have also been released [11-17]. In particular, Deepmind's Alphago [11, 12] is a representative algorithm that applies the deep neural network (DNN) to RL.

### B. Value-based RL versus Policy-based RL

The RL can be divided into value-based RL and policy-based RL, depending on whether the output of the DNN is state-value or policy. Value-based RL represents the RL with DNN, which approximates the state-value function, called Q-function. The agent chooses which action to take based on Q-function. In policy-based RL, action is not selected based on Q-function. However, the action is selected based on the policy, indicating the probability that which action will be taken at a given state. In other words, the policy is approximated through DNN. The output of the DNN represents a value between 0 and 1 since the policy is the probability of taking specific action in a state. Therefore, the activation function of the output layer of the DNN uses the softmax function. The reason is that the softmax function can be expressed as a probability distribution since the sum is "1". This policy-based RL has the advantage of converging quickly and being effective in a high-dimensional or continuous operating space.

### C. Deep Q Network

Deep Q Network (DQN) algorithm is a deep RL method that approximates a Q-function through a DNN rather than a Q-learning method obtained from a table format [13, 14]. The loss function of the DQN algorithm is given by

$$L_i(\theta_i) = \mathbb{E}_{s,a,r,s' \sim \mathcal{U}(\mathcal{D})}\Big[r + \gamma \max_{a'} Q(s',a';\theta_i^{-}) - Q(s,a;\theta_i)\Big] \quad (1)$$

where $L_i(\theta_i)$ is the learning loss, Q is a Q-value function and $\mathbb{E}_{s,a,r,s' \sim \mathcal{U}(\mathcal{D})}$ is the expected value. The s, a, r, $\gamma$ and $\theta_i$ represent the state, action, reward, depreciation rate, and weight parameter of the i-th term, respectively. The "'" symbol refers to the following state or action, and a transition is defined to be a quadruple $(s,a,r,s')$. The agent explores the environment and stores transitions in a buffer called 'replay memory ( $\mathcal{U}(\mathcal{D})$ )'. Correlation between transitions can interfere with learning and these transitions. It can be reduced since many transitions are randomly sampled from the stored replay memory. The DQN algorithm uses a target network to eliminate the correlation between transitions. The target network solves the instability caused by bootstrapping that creates another DNN for approximating the Q-function. The target network is updated from the DNN that approximates the Q-function and is fixed without being updated during the user-defined period. The Q-function of the target network is represented by $Q(s',a';\theta_i^{-})$.

### D. Proximal Policy Optimization

The trust region policy optimization (TRPO) [15] is an RL algorithm that avoids parameter updates that change the policy too much with a Kullback–Leibler (KL) divergence constraint on the size of the policy update at each iteration. For example, when we estimate the maximum point of function by using sampling, it will be difficult to find the global optimal point if the step size of the sampling is too large. Therefore, appropriate step size is required. In TRPO, a constraint is defined to learn in a reliable section. The degree of change in the policy $r_t(\theta)$ is given by

$$r_t(\theta) = \frac{\pi_\theta(a_t|s_t)}{\pi_{\theta old}(a_t|s_t)} \quad (2)$$

where $\theta$ is the DNN of TRPO, $\pi_\theta$ is the policy.
The loss function is given as

$$L(\theta) = maximize_\theta\ \widehat{E_t}[r_t(\theta)\widehat{A_t}] \quad (3)$$
$$subject\ to\ \widehat{E_t}[KL[\pi_{\theta old}(\cdot|s_t), \pi_\theta(\cdot|s_t)]] \leq \delta$$

where $\widehat{E_t}$ is expectation value and $\widehat{A_t}$ is the advantage function. In equation (3), the KL divergence of $\pi_{\theta old}$ and $\pi_\theta$ is constrained to $\delta$(pre-determined delta) or less to obtain $L(\theta)$. However, in order to learn this, a Hessian of KL divergence must be obtained to obtain $r_t(\theta)$, and in this process, a secondary differential is required. Therefore, it's inevitable to perform a large amount of computation.

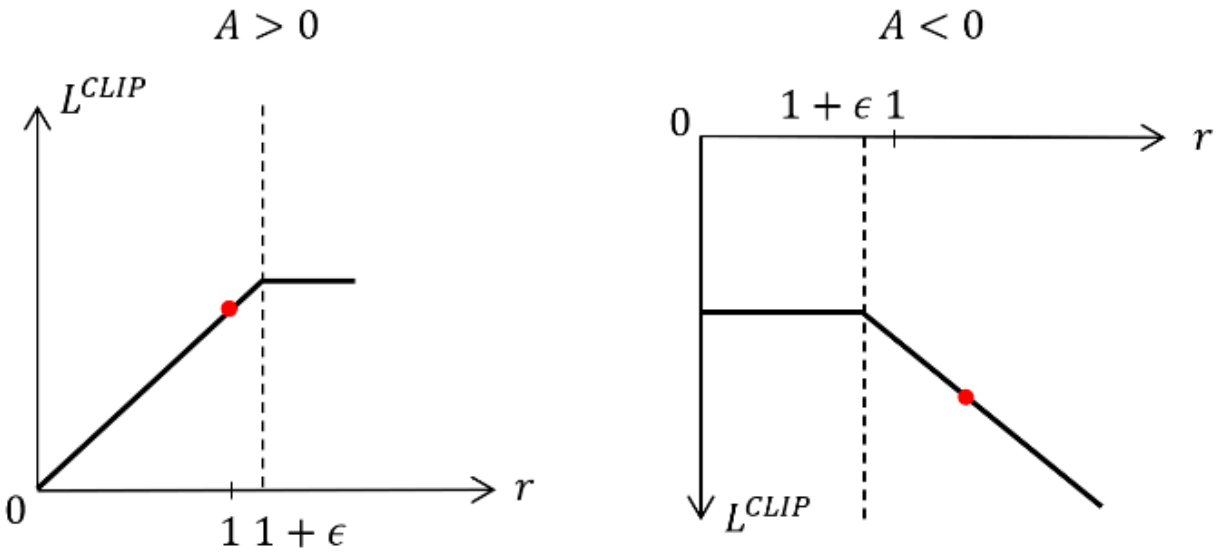

Fig. 1. Clipping of PPO

A method of solving the problem of the TRPO is the PPO [16]. Since it is complicated to obtain the Hessian matrix of the TRPO, the main concept of PPO is simply clipping the constraint. The Hessian matrix is a square matrix of second-order partial derivatives of a scalar-valued function. In this way, the Hessian matrix can be obtained only with the first-order differential equation. The loss of clipping $L^{CLIP}(\theta)$ is given by

$$L^{CLIP}(\theta) = \hat{E}_t\left[\min\left(r_t(\theta)\widehat{A_t}, clip(r_t(\theta), 1-\epsilon, 1+\epsilon)\widehat{A_t}\right)\right] \tag{4}$$

where $E_t$ is an expectation value and 'clip' is the clipping function like Fig. 1 and 'min' is the function that selects the minimal one and '$\epsilon$' is a hyperparameter that determines the amount of trim. From equation (4) and Fig. 1, it can be seen that $L^{CLIP}$ is obtained through clipping.

### *E. Transfer Learning*

The TL [18] focuses on storing knowledge gained while solving one problem and applying it to a different, however related problem. From a practical point of view, it has the potential to significantly improve the sample efficiency of RL by reusing information from previously learned problems for learning new problems.

## III. Proposed Method

In this section, a cleaning simulation environment and the proposed method are described. We create our custom cleaning simulation environment for this study. The proposed method featured by TL, DNUT, RS, and ES increases performance by changing the reward method, training with high-quality data, and providing additional information.

### *A. Cleaning Simulation Environment*

A simulation environment is created by using 'Tkinter' which is a Python GUI toolkit. The 5x5(25 pixels) and 20x20(400pixels) cleaning environments are used in this study. The cleaning robot is referred to as an agent. Fig. 2 shows the 5x5 simulation environment of the agent, which is used in this study. The red square is the cleaning robot that refers to the agent, and the white dot in the right center of the red square refers to the head of the agent. In the cleaning simulation environment, the yellow circles represent uncleaned areas, and the blue circles represent cleaned parts. When the simulation starts, all tiles are reset to a yellow circle, and if the agent cleans all tiles the simulation ends.

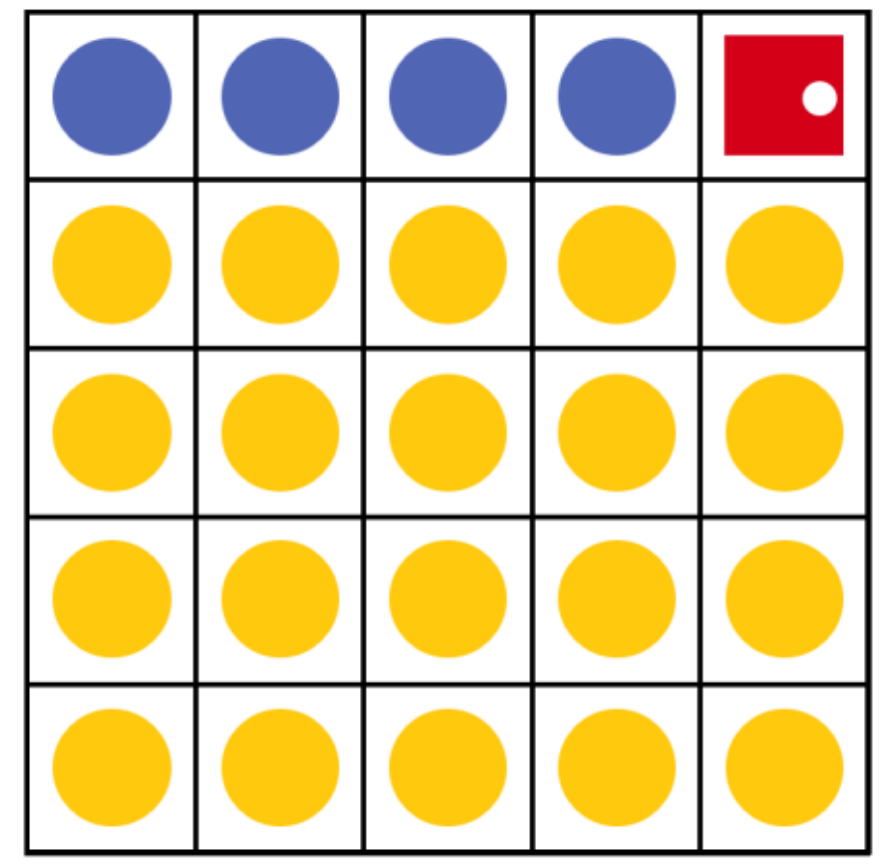

Fig. 2. 5x5 simulation environment for cleaning robot

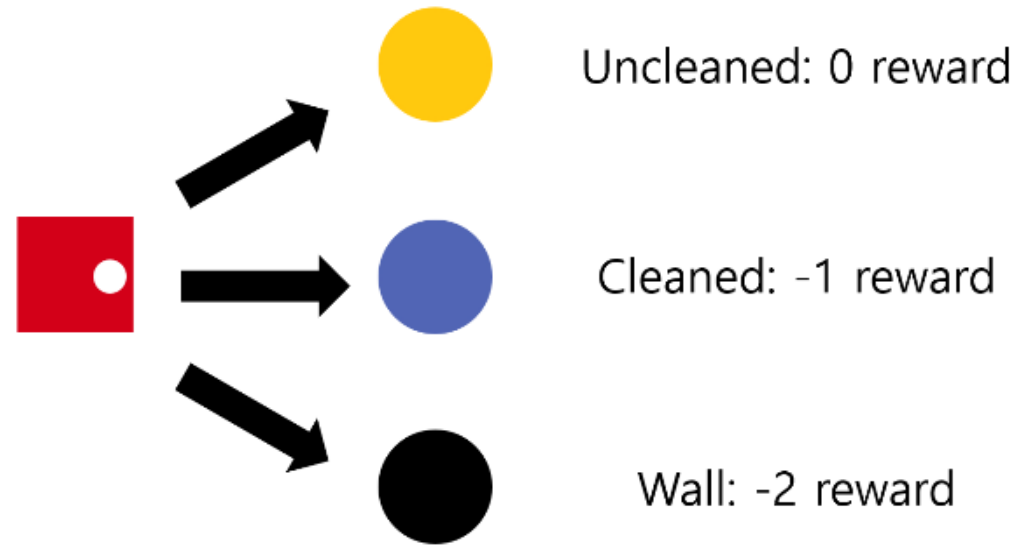

Fig. 3. Rewards of cleaning robot for each type of action

Fig. 3 is a schematic diagram of the reward of the action of the cleaning robot. If the agent goes to an uncleaned tile (yellow), it receives a reward of '0'. If the agent goes to cleaned tiles (blue), it receives a reward of '-1', and if the agent goes to a wall or obstacle (black), it receives a reward of '-2'. This system makes the agent move to uncleaned tiles by providing a negative (-) reward when the agent moves to cleaned or wall tiles.

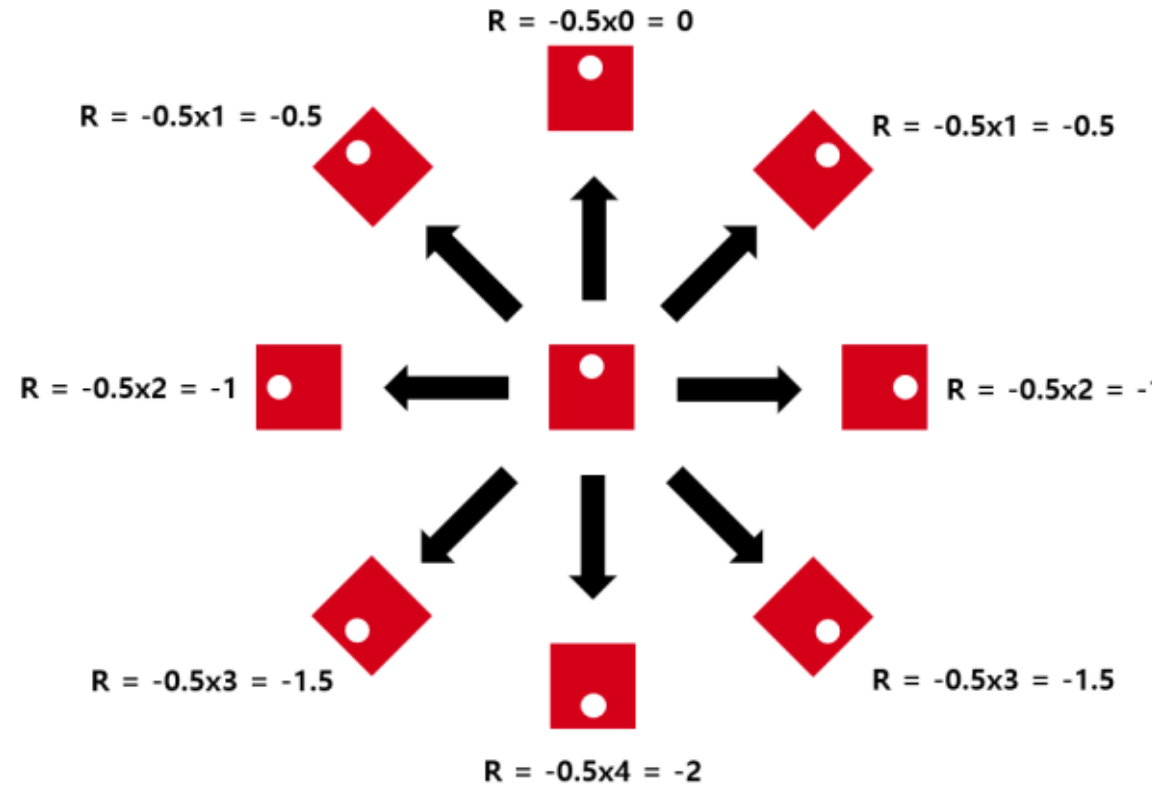

Fig. 4. Rewards of cleaning robot for each type of rotation

A rotation reward is additionally considered here. When the agent rotates, the position of the robot is unchanged. It keeps cleaning the same place until it has finished spinning since this agent is still running. Such a rotation also uses a lot of electricity. Therefore, in order to reduce the number of rotations as much as possible, a rotation reward is given as shown in Fig. 4. The agent is given a reward of -0.5 for each 45-degree rotation and assumed that the rotation is no more than 180 degrees since it can be clockwise or counterclockwise. That is, a minimum of -2 reward (180 degrees rotation) is received at a time. In addition to this, the agent can take one of the 8 actions in each step. It can only rotate in units of 45 degrees and cannot move at an angle less than that.

### *B. Proposed Method*

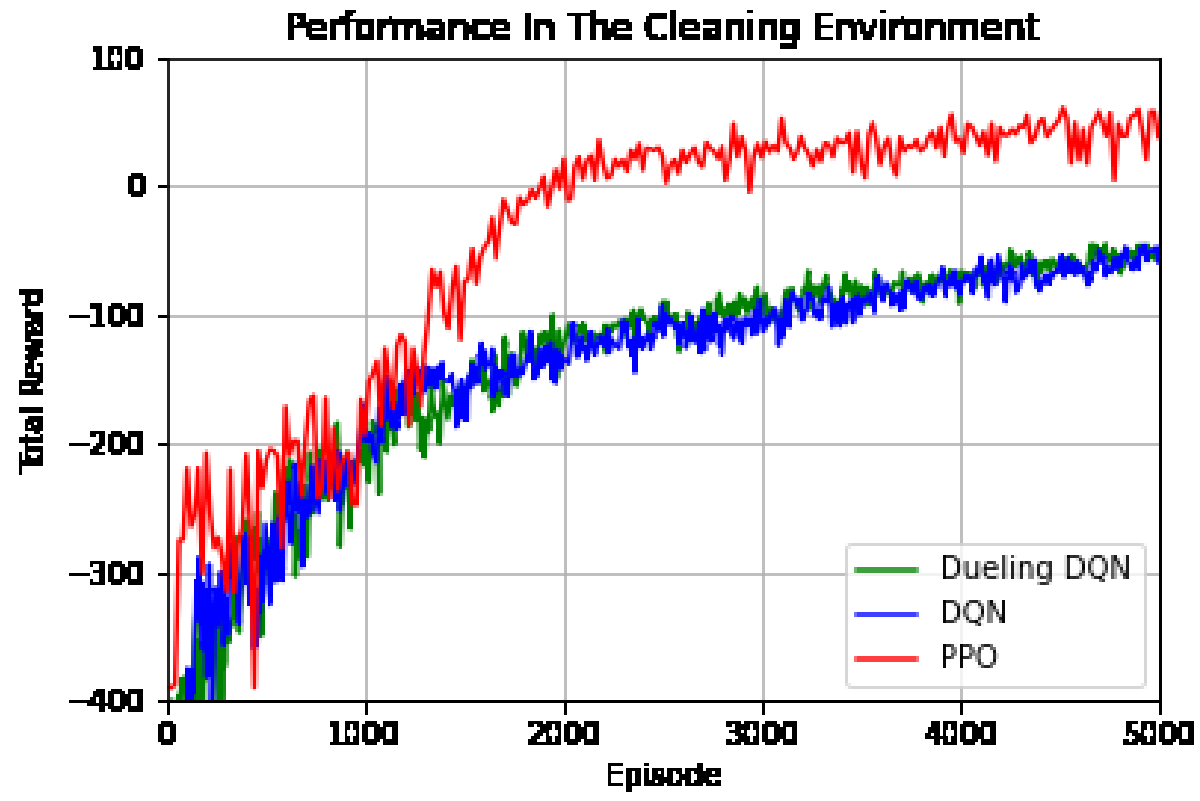

Fig. 5. Variation of total reward with PPO, DQN, Dueling DQN in the 5x5 cleaning environment

The PPO, DQN, and Dueling DQN are tested. As seen in Fig. 5, the PPO is the fastest converging algorithm in a cleaning environment. Therefore, it is judged that policy-based RL is more appropriate. To verify this, we tested the performance of various algorithms in the cleaning simulation environment. As a result, the PPO algorithm performs the best, as shown in Fig. 5. Therefore, this algorithm is adopted for further experiments.

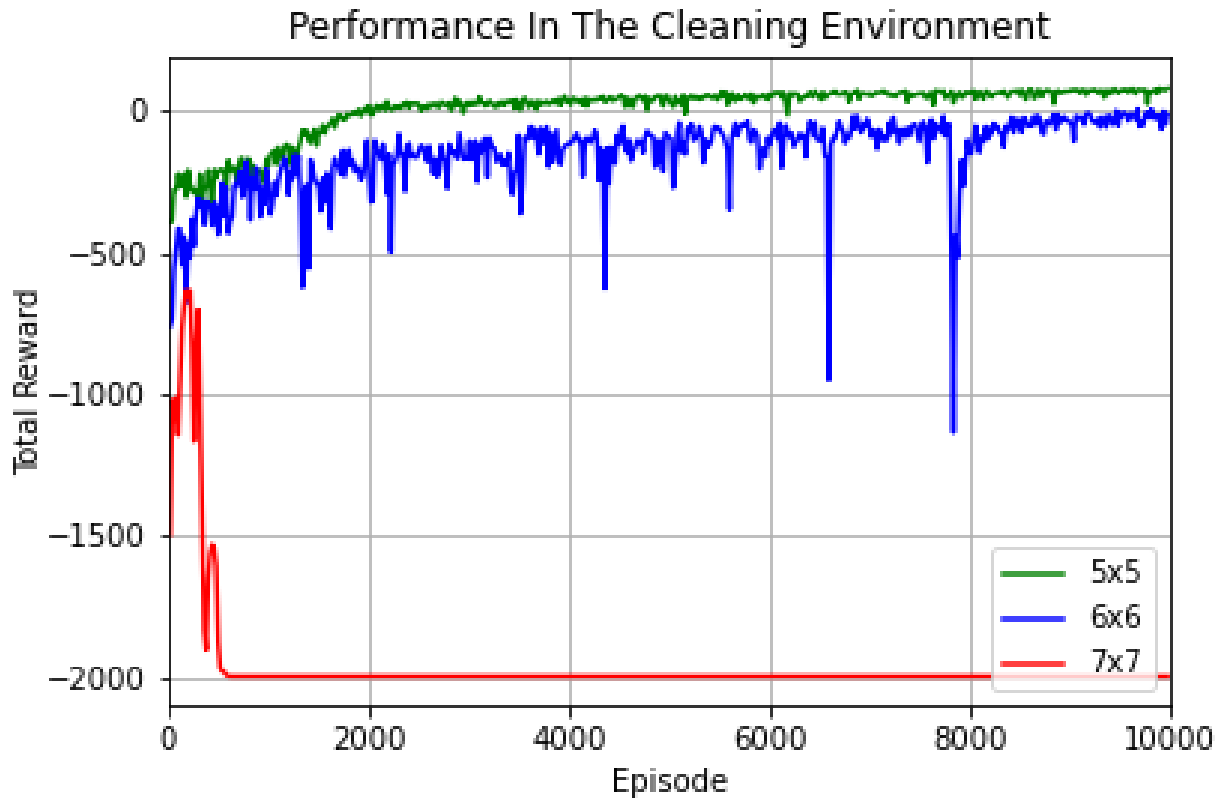

Fig. 6. Performance of PPO in various cleaning environments

Initially, coordinates (x, y) and rewards (r) of all tiles are set to the input state. Fig. 6 shows the total reward for each size of the cleaning environment. It can be seen that learning has not been performed as the size of the cleaning simulation environment increases. With the 7x7 environment, learning is not possible at all. Indeed, as the size of the simulation environment increases, the input state size also increases. For example, in the 5x5 environment, the input state size is $5 \times 5 \times (2 + 1) = 75$, however in the 7x7 environment, it almost doubles, i.e., $7 \times 7 \times (2 + 1) = 147$. Consequently, it becomes more difficult to train the DNN, and as a result learning is not possible. To solve this problem, different techniques are used. These techniques keep the input state size constant, regardless of the size of the cleaning environment. The techniques are described as follows.

1) Transfer learning in cleaning environment

The TL [19, 20] is a crucial foundation for learning in RL, which can apply the information acquired from previously completed tasks (known as source tasks) to a new task (called target task). Retrieving the necessary training data to recreate the models is either expensive or impossible in many real-world environments. Transfer of information or learning between task domains would be advantageous in these circumstances. The TL focuses on storing knowledge gained while solving one problem and applying it to a related problem.

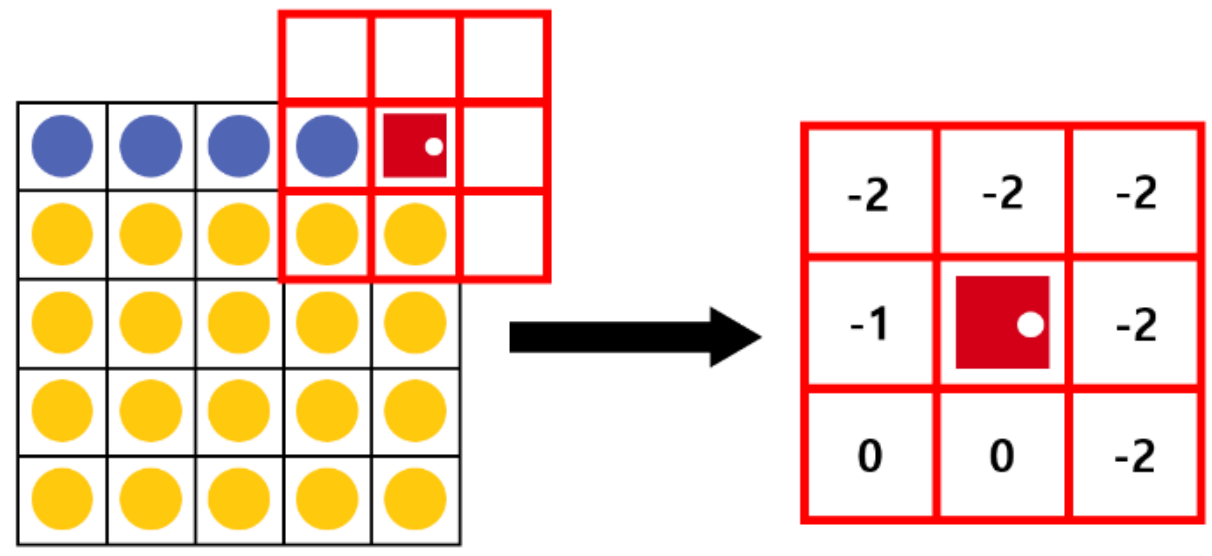

Fig. 7. Reward distribution around the location of the cleaning robot when the cleaning robot is at a corner

The method of detecting tile rewards around the agent and putting them in the input state is presented in Fig. 7. In this Fig., it can be seen that each of the eight boxes contains rewards of cleaned tiles, uncleaned tiles, and wall (obstacle) tiles. This allows the input state size to be fixed regardless of the size of the cleaning environment. The reward is notified to the agent when the next action is taken and the number of boxes is set to 8, which is equal to the number of actions, to minimize the input state size.

2) Detection of nearest uncleaned tile

The DNUT is a technique that the agent detects the nearest uncleaned tile. The agent can only see eight tiles around and has not much information to learn. If the surrounding area of the agent is already cleaned or there is an obstacle tile, the agent assumes that all of the cleaning environment's tiles are already cleaned and stops cleaning. To avoid this situation, the agent detects the closest uncleaned tile, and the relative location information of it is known to the agent. This allows the agent to clean without being trapped in a specific place, as shown in Fig. 8.

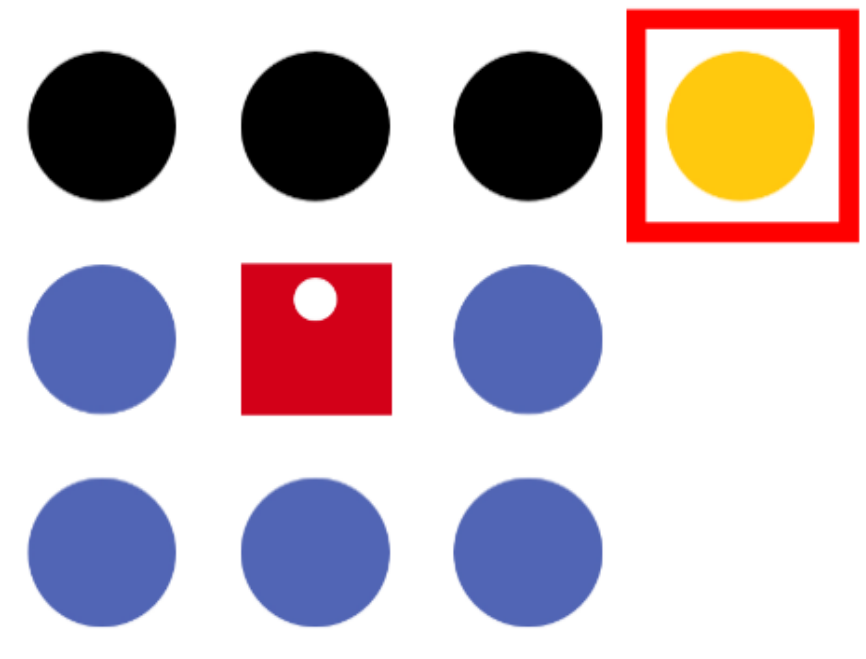

Fig. 8. Unclean tiles closest to the agent

3) Reward shaping combined with stacked value

The RS [21] is an effective technique that adjusts the reward method to make a problem easier to learn. This is a significant concept for improving the performance of learning. In this technique, we use a constant that is called ‘stacked value’. It indicates how well the agent is doing continuously. If the agent receives a positive (+) or zero reward (0), the stacked value is +1. If the agent receives a negative (-) reward, the stacked value is initialized to '0'. The additional ‘ $R^{\text{stacked value}}$ ’ reward is provided. As a result of the experiment, the learning performance is the best at R=1.5. For example, if the agent received +1 reward three times continuously, it receives an additional $1.5^1 + 1.5^2 + 1.5^3 = 1.5 + 2.25 + 3.375 = 7.125$ reward. Therefore, the total reward is $3 + 7.125 = +10.125$. If the agent received -1 reward three times continuously, it receives an additional $1.5^0 + 1.5^0 + 1.5^0 = 3$reward. Therefore, the total reward is $-3 + 3 = 0$. Hence, the agent tends to receive positive reward continuously, instead of receiving a negative reward. This can improve performance by additional reward to the agent.

4) Elite set

The Elite Set (ES) is a collection of high-performance data for effective learning [16, 22]. It prevents the agent from learning bad episodes. Due to the nature of the PPO algorithm, it does well when learning is good, however, sometimes it doesn't. Therefore, the agent is sometimes stuck in a specific area during learning. To minimize these situations, we introduce the concept of the ES. If the agent proceeds more than 500 steps in one episode, it is terminated and starts the next episode. It means, the agent improves performance by not learning the data of the bad performance episode.

## IV. Simulation Environment and Results

This section describes the experiment environment and presents the experimental results of the proposed method.

### A. *Simulation environment*

We trained the agent through the PPO algorithm in the 5x5 environment shown in Fig. 2. Five different algorithms that are without DNUT method, without RS method, without ES method, the original PPO algorithm, and the proposed method are employed for learning**.** The proposed models that are trained in a 5x5 environment are tested in the 20x20 environment shown in Fig. 11(a). The area of a tile is 0.5m × 0.5m = $0.25m^2$ and thus the total area is $100m^2$, which is a typical size of a living room.

### B. *Simulation result*

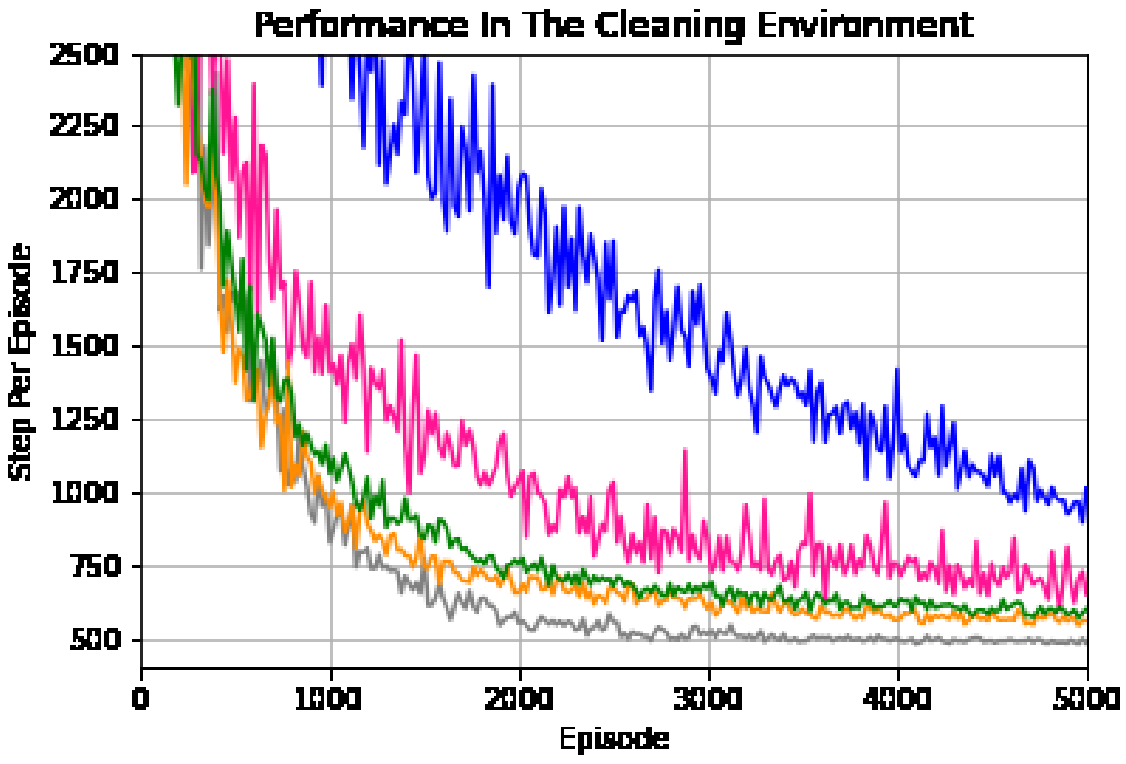

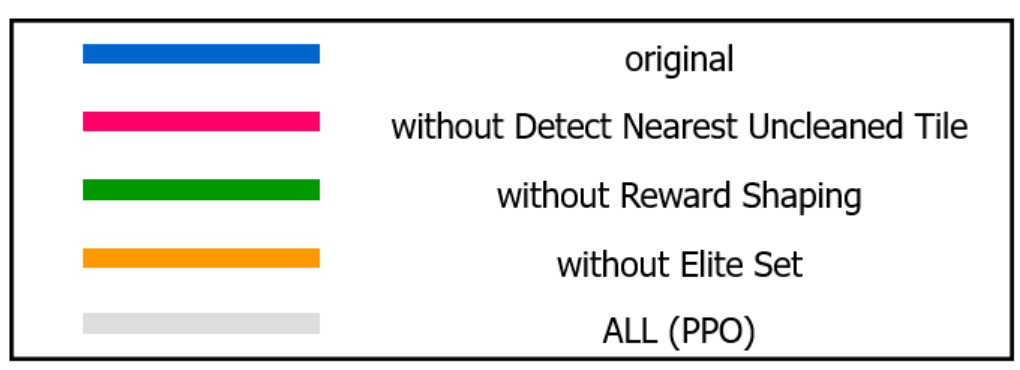

Fig. 9. Ablation study of proposed method denoted by ALL(PPO) and separated legend of Fig. 9

Fig. 9 shows the simulation results based on the techniques mentioned above. It shows the Ablation study of the proposed method in the 5x5 cleaning environment shown in Fig. 2. In Fig. 9, the blue one represents the performance of the original PPO algorithm, the pink one represents the algorithm without DNUT method, the green one is without RS, the orange one is without ES method, and finally, the gray one represents the performance of the proposed method. The x-axis indicates the number of episodes, and the y-axis represents the number of steps until the agent terminates the one episode. The smaller number of steps represents better performance. Among all the techniques in this study, the original technique, as shown in Fig. 9, leads to poor performance, while the proposed method demonstrates the best performance.

Results shown in the Fig. 11(b)-(e) are obtained by applying the PPO algorithm without DNUT, without RS, without making ES, with ALL(PPO)) trained with 10000 episodes in a 5x5 cleaning environment. Fig. 10 shows the comparative performance of the ALL (PPO) algorithm to other techniques and shows that the performance of the ALL (PPO) algorithm is much better than other techniques, e.g., PPO without DNUT, PPO without RS, PPO without making ES. In Fig, the score represents the total reward. When learning during the 10000 episodes, the ALL (PPO) algorithm, e.g., the proposed method, clean all tiles except 20 tiles in a 20x20 cleaning environment. When this algorithm is trained during 50000 episodes, it can clean all tiles as seen in Fig. 11(f).

We compare this proposed method with other conventional algorithms that are characterized by 'Random' and 'Zigzag' movements of the robot as shown in Table I. In Table I, the first one, the Random algorithm, is an algorithm that makes the robot move in a random direction when the obstacle appears while going straight as described in Section I. The second one, the Zigzag algorithm, makes the robot move in a zigzag path, which is commonly known to be the most efficient cleaning behavior in these day's markets. The test environment of the proposed method is 20x20 as shown in Fig. 11(a). The agent is given an objective reward based on the

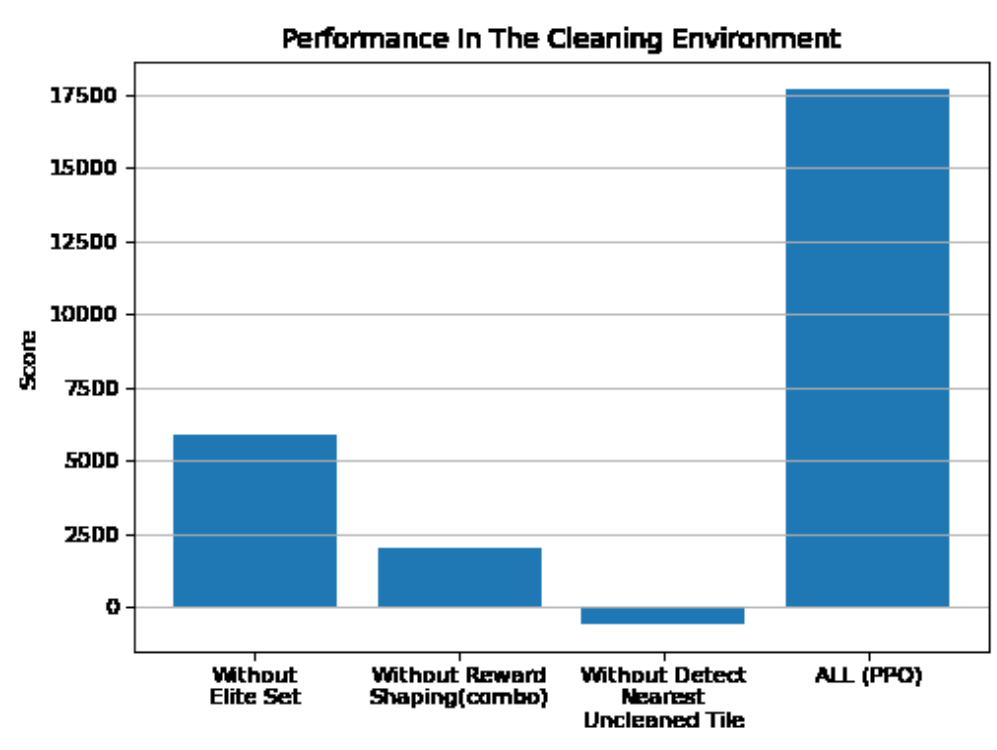

Fig. 10. Score obtained in 20x20 cleaning environment

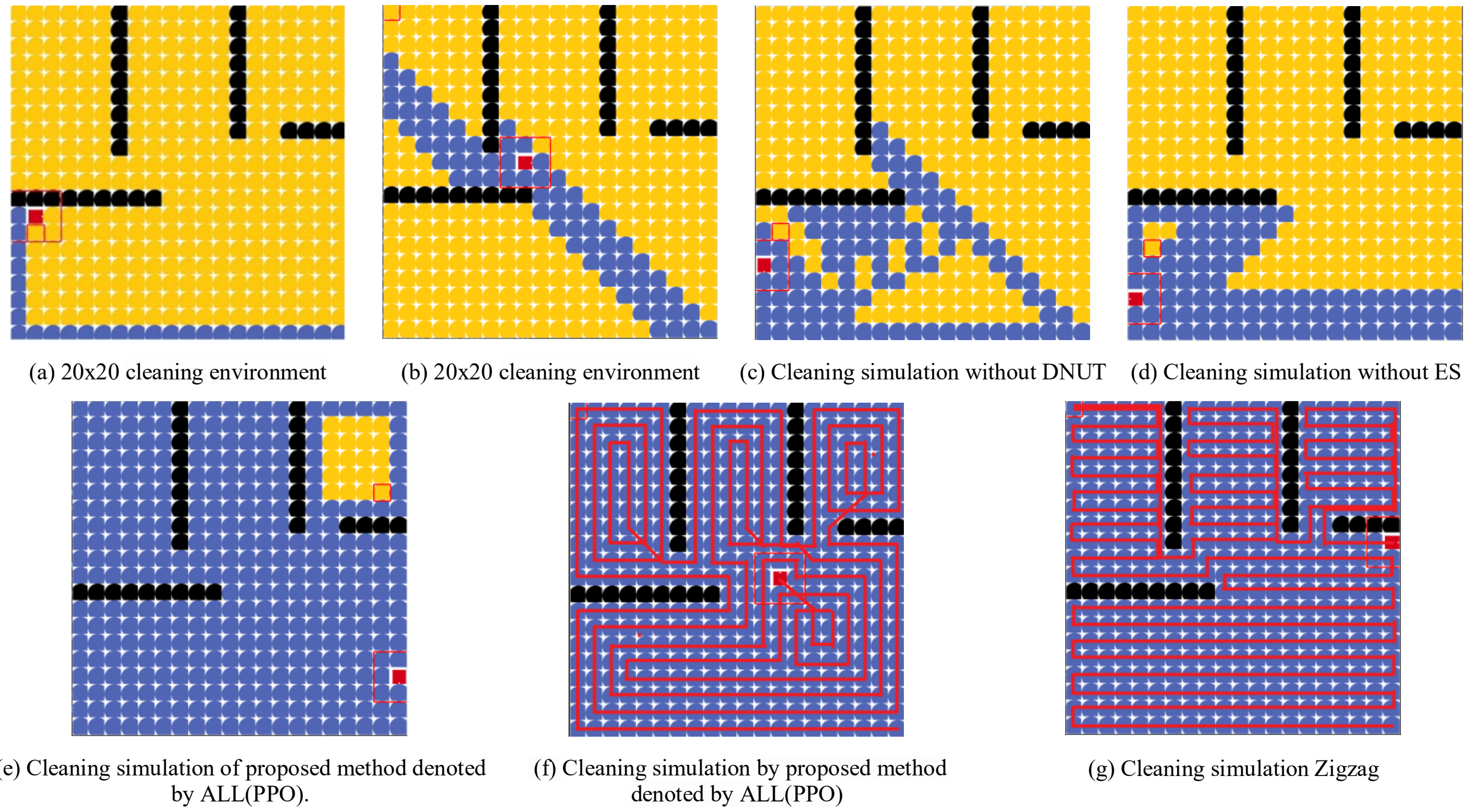
(a) 20x20 cleaning environment (b) 20x20 cleaning environment (c) Cleaning simulation without DNUT (d) Cleaning simulation without ES

(e) Cleaning simulation of proposed method denoted by ALL(PPO). (f) Cleaning simulation by proposed method denoted by ALL(PPO) (g) Cleaning simulation Zigzag distance traveled and the number of rotations until the entire room is cleaned. This allows a more fair comparison of performances. The distance reward is an actual length(m) that the agent has traveled in the real world and the degree reward is given 1 for each 45 degrees rotation of the agent.

Fig. 11. Results obtained from the proposed method in 20x20 cleaning environment

TABLE I. COMPARATIVE PERFORMANCE MEASURED BY TRAVEL DISTANCE AND ROTATION BY THREE MODELS RANDOM, ZIGZAG, ALL (PPO) MODELS

| **Table Head** | **Model** | | |
|---|---|---|---|
| | ***Random*** | ***Zigzag*** | ***ALL(PPO)*** |
| Distance(m) | 33830.1 | 201.0 | 194.0 |
| Rotation | 24847 | 172 | 162 |

As seen in Table I, the ALL (PPO) algorithm performs 17438.20% better than the random algorithm and 3.61% better than the Zigzag algorithm in terms of travel distance. In addition, in terms of rotation, it is 15337.65% better than the Random algorithm and 6.17% better than the Zigzag algorithm.

## V. CONCLUSION

In this paper, the PPO algorithm, an on-policy algorithm among RL algorithms fit to optimization of the path of cleaning robots is combined with other supplementary techniques to provide improved performance of cleaning robot. The agent achieves superior performance through the TL technique in an environment where learning is slow or not done with the existing algorithms. The proposed method shows the performance achieved without the training of the agent in a specific environment. The proposed method performs 17438.20% better than the 'random' algorithm and 3.61% better than the 'zigzag' algorithm in terms of travel distance. In addition, in terms of rotation, it is 15337.65% better than the 'random' algorithm and 6.17% better than the 'zigzag' algorithm.

## ACKNOWLEDGMENT

This work was supported by the Institute for Information communications Technology Promotion (IITP) grant funded by the Korean government (MSIT) (No.2020-0-00440, Development of Artificial Intelligence Technology that continuously improves itself as the situation changes in the real world).